\documentclass[review]{elsarticle}

\usepackage{lineno,hyperref}
\modulolinenumbers[5]

\journal{Journal of Pattern Recognition}

\usepackage[numbers]{natbib}
\usepackage{framed, multirow, array}
\usepackage{graphicx}
\usepackage{threeparttable}
\usepackage{color,soul}
\usepackage{caption}
\usepackage{picins}
\usepackage{amssymb}
\usepackage{amsmath}
\usepackage{latexsym}
\usepackage{pifont}
\usepackage{tabularx,arydshln}
\usepackage{url}
\usepackage{xcolor}
\usepackage{subfigure}
\usepackage{multicol}
\definecolor{newcolor}{rgb}{.8,.349,.1}
\newcommand{\xmark}{\ding{55}}%
\newcommand{\cmark}{\ding{51}}%

\begin{document}





    
    
    

    

\title {Non-Volume Preserving-based Fusion to Group-Level Emotion Recognition on Crowd Videos}      

\author[1]{Kha Gia Quach} 
\author[2]{Ngan Le}
\author[1]{Chi Nhan Duong}
\author[2]{Ibsa Jalata}
\author[3]{Kaushik Roy}
\author[2]{Khoa Luu}

\address[1]{Computer Science and Software Engineering, Concordia University, Canada}
\address[2]{Computer Science and Computer Engineering, University of Arkansas, USA}
\address[3]{Department of Computer Science, North Carolina A\&T State University, USA}

\begin{abstract}
{\small
Group-level emotion recognition (ER) is a growing research area as the demands for assessing crowds of all sizes are becoming an interest in both the security arena as well as social media. This work extends the earlier ER investigations, which focused on either group-level ER on single images or within a video, by fully investigating \textbf{group-level expression recognition on crowd videos}.
In this paper, we propose an effective \textit{deep feature level fusion mechanism} to model the spatial-temporal information in the crowd videos. 
In our approach, the fusing process is performed on the deep feature domain by a generative probabilistic model, Non-Volume Preserving Fusion (NVPF), that models spatial information relationships. 
Furthermore, we extend our proposed spatial NVPF approach to the spatial-temporal NVPF approach to learn the temporal information between frames. 
To demonstrate the robustness and effectiveness of each component in the proposed approach, three experiments were conducted: (i) evaluation on AffectNet database to benchmark the proposed EmoNet for recognizing facial expression; (ii) evaluation on EmotiW2018 to benchmark the proposed deep feature level fusion mechanism NVPF; and, (iii) examine the proposed TNVPF on an innovative Group-level Emotion on Crowd Videos (GECV) dataset composed of 627 videos collected from publicly available sources. GECV dataset is a collection of videos 
containing crowds of people. Each video is labeled with emotion categories at three levels: individual faces, group of people, and the entire video frame.}
\end{abstract}

\maketitle

\section{Introduction}

\begin{figure*}[!t]
	\centering 
	\begin{subfigure}
    \centering
    \includegraphics[width=0.32\columnwidth]{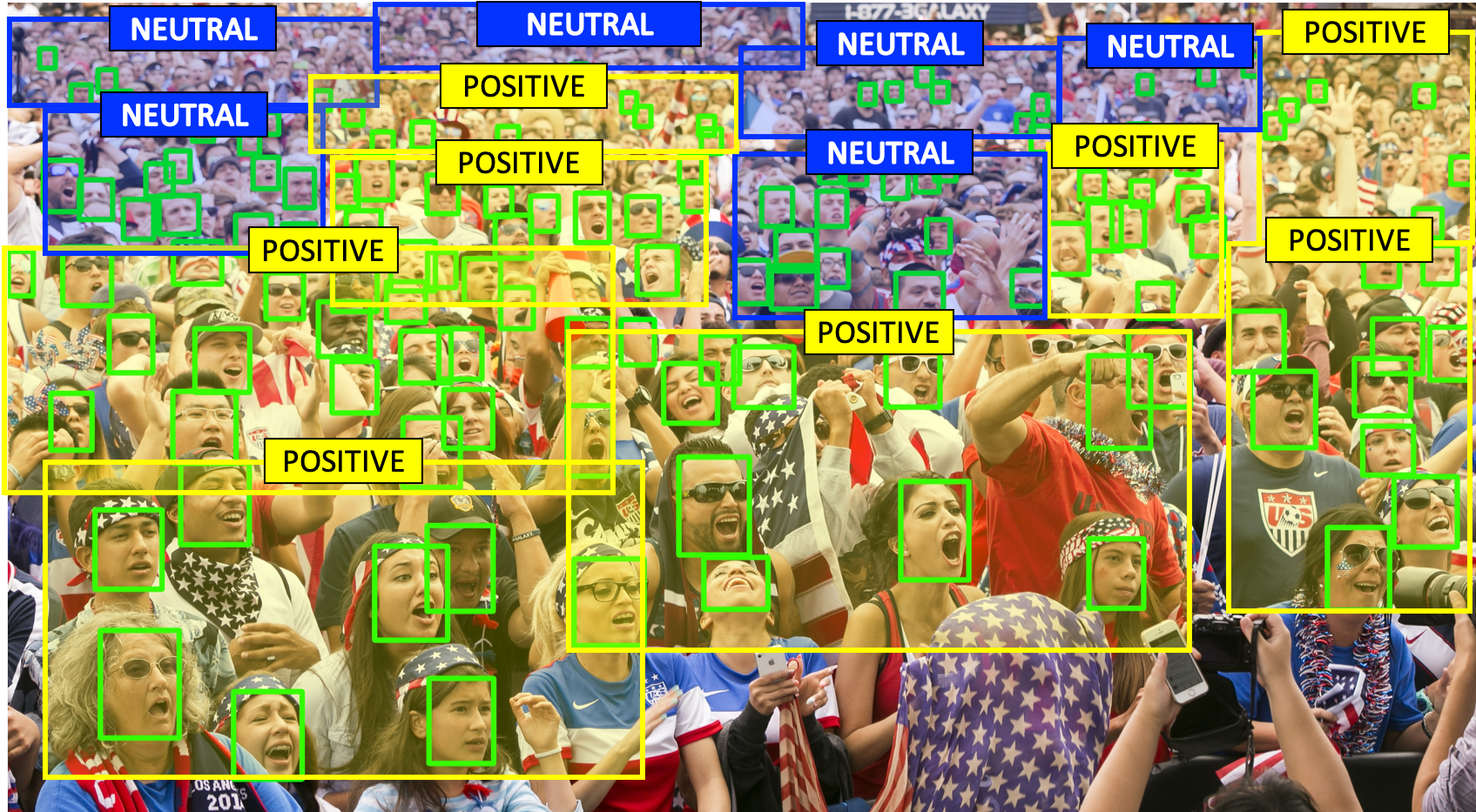}	
    \end{subfigure}%
    \begin{subfigure}
    \centering
    \includegraphics[width=0.32\columnwidth]{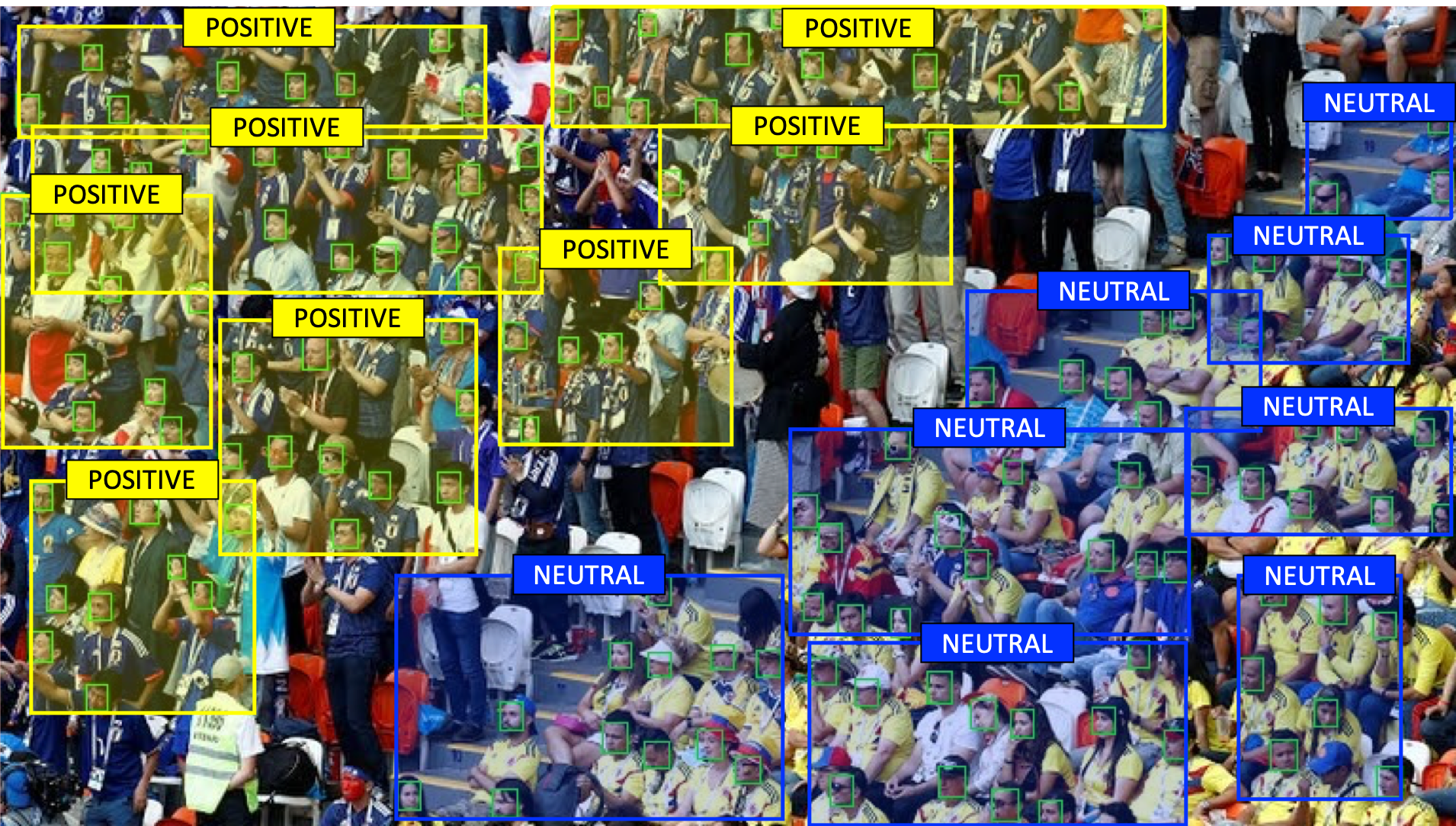}	
    \end{subfigure}%
	\begin{subfigure}
	\centering 
    \includegraphics[width=0.32\columnwidth]{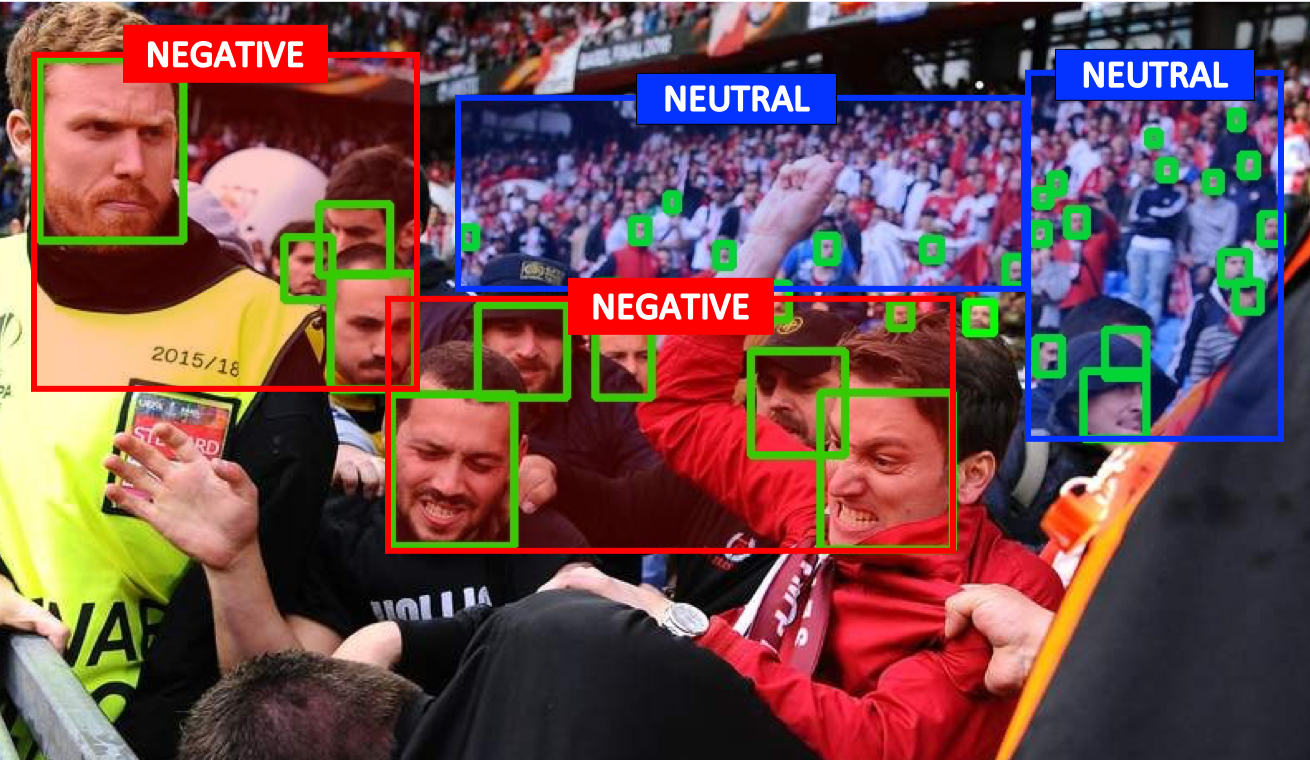}
    \end{subfigure}%
    \caption{ Samples of Group-level ER on Crowd Videos with manually annotated labels. Multiple emotions present within an image and human faces are unclear at further distance. \textbf{Best viewed in color.}}
   \label{fig:fig2}
\end{figure*}

Emotion recognition (ER) based on human's facial expression via facial action units (FACS), i.e. movement of facial muscles, has been studied for years in the field of affective computing, e-learning, health care, virtual reality entertainment,  and human-computer interaction (HCI). ER approaches can be technically categorized into two groups: (i) \textit{Individual ER}, (ii) \textit{Group-level ER}. 
While the studies in individual ER are quite mature, the research in group-level ER is still in its infancy. A challenge of group-level ER is the detection of all faces in the group and aggregating the emotional content of the group across the scene (image or video) as shown in Figure \ref{fig:fig2}. 

Traditional approaches to ER are based on hand-designed features as illustrated by \citep{Shan2009, Kahou2015}. However, with the emergence of deep learning, copious large-scale datasets, and the compute power of graphical processors, computer vision tasks have seen enormous performance gains, this is indeed true for individual (traditional) ER.  Compared to traditional hand-crafted models, an optimal deep learning model is capable of extracting deeper discriminate features. These deep feature-based ER solutions have proven capable of not only classifying group-level emotions on single images (see Sec. \ref{subsec:GroupEmot}), but videos for individual or group ER (see Sec. \ref{subsec:EmotVid} and \ref{subsec:Group-EmotVid}). 

Unlike prior work tackle ER on videos, 
this work examines group (crowd) ER responses by categorizing at different levels from bottom to top, i.e. individuals, group, and the whole video, as positive, negative, or neutral, except for individuals with eight emotion categories.  
Furthermore, a new fusion approach to facial feature-based group-level ER has been developed over the simplified approaches presented to date in which the final decision is based on the group of faces as represented by some form of averaging or winner take all voting paradigm. 

This work introduces a new deep feature-based fusion mechanism termed Non-volume Preserving Fusion (NVPF) which is demonstrated to better model the spatial relationship between facial emotions among the group within an image or still frame. In addition to the proposed NVPF mechanism, we solve the crowd problem in which multiple emotions are presented. On top of that, this mechanism is a remedy for unclear emotion due to the resolution of the face--the face is too small to register any emotion as shown in Figure \ref{fig:fig2}.
The contribution of our proposed deep feature-level fusion approach to group-level ER on crowd videos can be summarized as follows:
\begin{itemize}
    \item To the best of our knowledge, this is one of the initial works to fully address \textit{group-level emotion on crowd videos} with multiple emotions across the crowd in videos at variable face resolution.
    \item Propose a high-performance and low-cost deep network for facial expression recognition named \textit{EmoNet} to robustly extract facial expression features for individuals.
    \item Present a novel deep learning-based fusion mechanism named \textit{Non-volume Preserving Fusion} (NVPF) to model the \textit{feature-level} spatial relationship between facial expressions within a group.
    \item The presented framework is then extended in a new \textit{deep network Temporal Non-volume Preserving Fusion} (TNVPF) to tackle the temporal-spatial fusion mechanism on videos.
    \item Differentiated from previous work that only presents one emotion status for the entire image, the proposed method can \textit{cluster multiple emotion regions} in images or videos as given in Fig.\ref{fig:fig2}.
    \item Finally, a \textit{new dataset} GECV is introduced for the problem of group-level ER on crowd videos.
\end{itemize}

\begin{table*}[h]
\caption{Comparisons on facial feature-based expression recognition between our and other recent  methods, where \xmark $ \: $ represents \textit{unknown} or \textit{not directly applicable} properties. Note: Long Short-Term Memory (LSTM), Random Forest (RF), Support Vector Machine (SVM), Standard Deviation (STD).}
\label{tab:MethodSumm}
\centering
\resizebox{1.01\textwidth}{!}{
\begin{tabular} {l|c|c||c|c|c||c|c} \hline

 & \begin{tabular}[c]{@{}c@{}} {\textbf{Feature}} \\ {\textbf{Fusion}} \end{tabular}  &  \begin{tabular}[c]{@{}c@{}} {\textbf{Fusion}} \\ {\textbf{Mechanism}} \end{tabular}  &  \begin{tabular}[c]{@{}c@{}} {\textbf{Prediction}} \\ {\textbf{Level}} \end{tabular} &  \begin{tabular}[c]{@{}c@{}} {\textbf{Very}} \\ {\textbf{Crowded}} \end{tabular} & \begin{tabular}[c]{@{}c@{}} {\textbf{Unclear}} \\ {\textbf{Faces}} \end{tabular} & 
 \textbf{Modality}  
  &  \textbf{Classifier}  \\
 \hline  \hline 

 Wei et al. \citeyear{wei2017new}  & \cmark & LSTM & Group & \xmark & \xmark & Image & SVM \\  
 Tan et al.  \citeyear{tan2017group} & \xmark & Average & Group & \xmark & \xmark & Image & Softmax \\  
 Rassadin et al. \citeyear{rassadin2017group} & \cmark & Median & Group & \xmark & \xmark & Image & Softmax \\  
 Gupta et al. \citeyear{Gupta2018attention} & \cmark & Average & Group & \xmark & \xmark & Image & Softmax \\  
 Khan et al. \citeyear{Khan_2018} & \xmark & Average & Group & \xmark & \xmark  & Image & RF \\ 
 \hline
 Hu et al. \citeyear{hu2017learning}  & \xmark & Concatenate & Individual & \xmark & \xmark & Video & Softmax \\
 Knyazev et al. \citeyear{knyazev2017convolutional} & \cmark  & Mean, STD & Individual & \xmark & \xmark & Video & SVM \\ 
 Fan et al. \citeyear{fan2018video} & \cmark & Activation Sum & Individual & \xmark & \xmark & Video & Softmax \\  
  \hline
 Favaretto et al. \citeyear{favaretto2019detecting} & \xmark  & mapping & Group & \xmark & \xmark & Video & \xmark \\
 \hline  \hline 
 \textbf{Ours (NVPF / TNVPF)} & \cmark & \textbf{NVPF} & \begin{tabular}[c]{@{}c@{}} {\textbf{Multiple}} \\ {\textbf{(Individual/Group)}} \end{tabular} & \cmark & \cmark & \textbf{Image/Video} & \begin{tabular}[c]{@{}c@{}} {\textbf{Softmax}} \\ {\textbf{(on NVPF)}} \end{tabular} \\
 \hline
 \hline
\end{tabular}
}
\end{table*}

\section{Related Work} 
\label{sec:Relwork}

In this section, we review the recent work on group-level ER on images (sub-section \ref{subsec:GroupEmot}) and individual ER on Videos (sub-section \ref{subsec:EmotVid}). Table \ref{tab:MethodSumm} shows the comparison in terms of facial feature-based expression recognition between our proposed framework and other state-of-the-art methods.

\subsection{Group-level ER on Still Image}
\label{subsec:GroupEmot}

Previous works on this task \citep{ 
dhall2015automatic, huang2015riesz}     
have focused on extracting scene features from the entire image as a global representation and facial features from faces in the given image as a local representation \citep{Luu_IJCB2011, Duong_ICASSP2011, Luu_BTAS2009, Luu_ROBUST2008}. 
In particular, one can observe the basic ideas among the EmotiW2018 group-based ER sub-challenge winners \citep{guo2017group, Gupta2018attention, Khan_2018, wang2018cascade} was to propose hybrid networks based on faces, scenes, skeletons, body, and visual attention to recognize group emotion. 
Most state-of-the-art approaches use "naive" mechanisms such as averaging \citep{Abbas_group, tan2017group, wang2018cascade}, concatenating \citep{wei2017new}, weighting \citep{guo2017group, tan2017group, rassadin2017group}, etc. to merge the global information and local representation. The averaging scheme, as identified in the work referenced above, is nothing more than a voting or majority selecting scheme. Concatenating or weighting introduced by Guo et al. (\citeyear{guo2017group}) utilized seven different CNNs-based models which have been trained on different parts of the scene, background, faces, and skeletons, which are optimized over the predictions. Tan et al. (\citeyear{tan2017group}) built three CNNs models for aligned faces, non-aligned faces, and entire images, respectively \citep{Luu_FG2011, Chen_FG2011, Luu_CAI2011, Le_JPR2015, Xu_TIP2015, Xu_IJCB2011}. Each CNN  produces scores across each class which is then combined via an averaging strategy to obtain the final class score. By contrast, Wei et al. (\citeyear{wei2017new}) modeled the spatial relationship between faces with an LSTM network. The local information of each face is presented by VGGFace-lstm and DCNN-lstm  while the global information is extracted by Pyramid Histogram of Oriented Gradients (PHOG), CENTRIST, DCNN, and VGG features. The local and global features are fused by score fusion. Rassadin et al. (\citeyear{rassadin2017group}) approach involved extracting feature vectors of detected faces using CNNs trained for face identification task. Random Forest classifiers were employed to predict the emotion score. Wang et al. (\citeyear{wang2018cascade}) proposed to use three types of hints, namely face, body, and global image, with three CNNs for each cue, then the final score is obtained by averaging all the scores from all faces, bodies, and global image. 

Different from other works on score fusion by weighting or averaging, Abbas et al. (\citeyear{Abbas_group}) utilized a densely connected network to merge 1x3 score vector from the scene and 1x3 score vector from the facial feature. 
Gupta et al. (\citeyear{Gupta2018attention}) proposed different weighted fusion mechanisms for both local and global information. Their attention model is performed at either feature level or score level. Applying ResNet-18 and ResNet-34 on both small face and big face was proposed by Khan et al. (\citeyear{Khan_2018}), which was designed as a four-stream hybrid network. 

In addition to recognizing the emotion of a group of people, \textit{Group Cohesion} (\citep{guo2019exploring, xuan2019group, zhu2019automatic}), i.e. the tendency for a group to be united for a common goal or emotion can be predicted.
In the EmotiW 2019 Challenge, the organizers presented their study on group cohesion prediction in static images \citep{ghosh2019predicting, dhall2019emotiw}. They extended the Group Affect (GAF) Database \citep{dhall2017individual} with group cohesion labels and introduced the new GAF Cohesion database. In their paper, they extracted image-level (global) and face-level (local) features using Inception V3\citep{szegedy2016rethinking} and CapsNet\citep{sabour2017dynamic}, respectively, to predict group cohesion. 
Recently, Mou et al. \citep{mou2019alone} proposed a framework to predict contextual information from individuals and groups with different settings, i.e. using both face and body behavioral cues, using multi-modal fusion and temporal modeling from videos by Long Short-Term Memory Networks (LSTM).
More recently, the emergence of large-scale datasets for crowd counting and localization, e.g. NWPU-Crowd \citep{wang2020nwpu}, etc. helps to push forward crowd scene understanding as seen in \citep{wang2020pixel} and \citep{wang2018detecting} proposed by Wang et al.

\subsection{Individual-level ER on Videos}
\label{subsec:EmotVid}

Kahou et al. (\citeyear{Kahou2013}) combined multiple deep neural networks including deep CNN, deep belief net, deep autoencoder, and shallow network for different data modalities on the EmotiW2013. This approach won the competition.  The temporal information between frames is fused through averaging the score decisions. A year later, the winner of EmotiW2014, Liu et al. (\citeyear{Liu2014}) used three types of image set models i.e. linear subspace, co-variance matrix, and Gaussian distribution; and three classifiers i.e. logistic regression, and partial least squares are investigated on the video sets. Similar to the work in \citep{Kahou2013}, the temporal information between frames is fusing through averaging on score decisions in \citep{Liu2014}. Instead of averaging, \citep{EbrahimiKahou2015} - winner of EmotiW 2015 utilized RNNs to model the temporal information. In this approach, Multilayer Perceptron (MLP) with separate hidden layers  for  each  modality which then are concatenated. Li et al. (\citeyear{li2019bi}) proposed a framework to predict emotion using two flows of information from a video, i.e. image and audio. For image flow, CNN-based networks were employed to extract Spatio-temporal features from both cropped faces and sequence of images. For audio flow, audio features, i.e. low-level descriptor and spectrogram, were extracted to compute a fused audio score. Finally, all scores were combined by weight summation.

Bargal et al. (\citeyear{Bargal2016}) used a spatial approach to video classification where the feature encoding module based on Signed Square Root (SSR) and $l_2$ normalization by concatenating FC5 of VGG13+FC7 of VGG16+pool of ResNet, and finally a Support Vector Machine (SVM) classification module. Fan et al. (\citeyear{fan2016video}) present a video-based ER system whose core module of this system is a hybrid network that combines RNNs and 3D CNNs. The 3D CNNs encode appearance and motion information in different ways whereas the RNNs encode the motion later. Hu et al. (\citeyear{hu2017learning}) presented Supervised Scoring Ensemble (SSE) by adding supervision not only to deep layers but also to intermediate and shallow layers. A new fusion structure, where class-wise scoring activation at diverse complementary feature layers are concatenated, is further used as the inputs for second-level supervision. This acts as a deep feature ensemble within a single CNN architecture. Recently, Later, Wang et al. (\citeyear{wang2019bootstrap}) proposed to extract multi-modal features, i.e. eye gaze, head pose, body posture, and action features, from a sequence of images using OpenFace \citep{amos2016openface}, OpenPose \citep{simon2017hand}, and Convolution3D \citep{tran2015learning}, and then ensemble these models by average weights. 
Recently, Sheng and Li (\citeyear{SHENG2021107868}) proposed a multi-task network to recognize identity and emotion from gait simultaneously.

\subsection{Group-level ER on Videos}
\label{subsec:Group-EmotVid}

Meanwhile, there are few studies on the analysis of crowd or analysis for violent behavior, e.g. Favaretto et al. (\citeyear{favaretto2019detecting} present a method to predict the personality and emotion of crows in videos. They detected and tracked each person and then recognized and categorized the Big Five Dimensions of Personality (OCEAN dimensions) and emotion in videos based on OCC emotion models.

From the aforementioned literature review, most of the prior work only tackled both problems of group-level ER and ER on videos via simple ensemble/fusion approaches. Furthermore, most of the previous work which makes use of facial-based feature can neither handle the cases when human faces have multiple resolutions nor deal with the scenario where multiple group emotions exist within an image. For example, crowd images/videos contains many human faces captured in a small portion (low resolution). As reviewed by Zhao et al. (\citeyear{zhao_tpami2021}), group emotion is a promising research direction for Affective Image Content Analysis.

\begin{figure*}[!t]
	\centering \includegraphics[width=1.0\columnwidth]{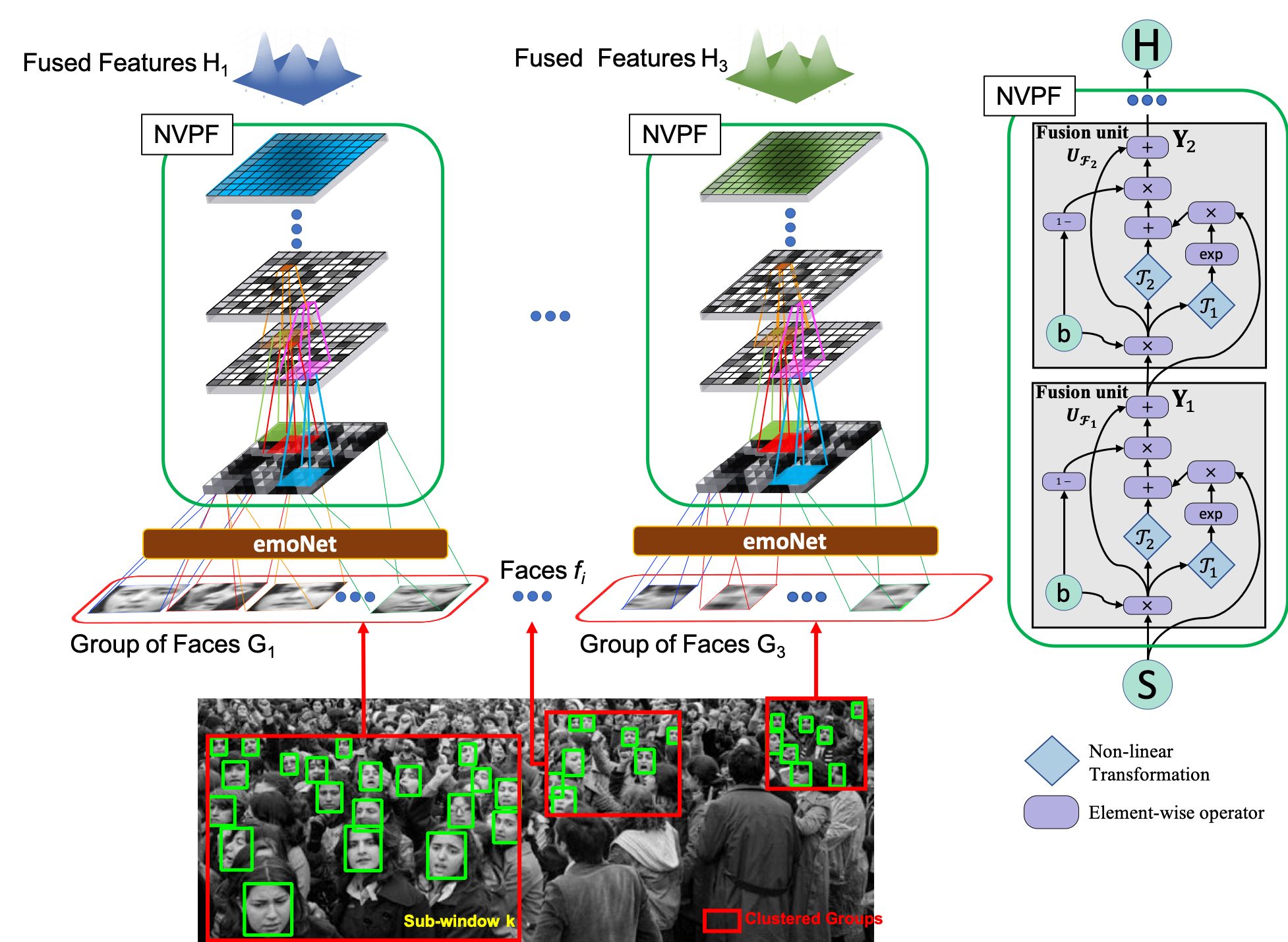}
    \caption{Illustrated the proposed framework on a single image with two components: (i) CNNs-based feature extraction EmoNet and (ii) Emotion fusion NVPF on detected face sub-windows} 
    \label{fig:overall_network_architecture}
\end{figure*}

\section{Our Proposed Approach}
In this section, we describe our proposed deep learning based approach to handle the problem of \textit{group-level ER on crowd videos in the wild}. Our proposed framework contains three components corresponding to (i) our new designed CNN framework named EmoNet to extract facial emotion features of a single facial image, (ii) a novel Non-Volume Preserving Fusion (NVPF) mechanism to model spatial representation between groups of multiple faces and, (iii) Temporal relationship embedding with Temporal NVPF structure, called TNVP to model the temporal relationship between multiple video frames. The first and second components are to learn the spatial representation of groups of people in a single image and it is equivalent to group-level ER on a still image. Figure \ref{fig:overall_network_architecture} shows the overall structure of our group-level ER on a still image. While the first component is to learn visual representation at individual-level, the second component makes use of NVPF approach is to extract visual representation at group-level and it can handle fusion faces at various resolutions. Then the temporal relationship between video frames (which are still images) is further exploited in the third component and its structure is presented in Figure \ref{fig:framework_videos} through Temporal Non-volume Preserving Fusion (TNVPF). 

In our proposed framework, detected faces are also clustered into groups based on relative spatial distance. Then, for each clustered group of faces, the extracted deep features are vectorized and structured as inputs to NVPF module to obtain the group-level facial expression features. 
Finally, the fused features for each frame and the whole video can be obtained by a temporal-spatial fusion approach, named Temporal NVPF, which can fuse and propagate features from frame to frame.
In addition to fused features at each level, those features can be used to provide predicted emotion categories at a specific level, i.e. individual faces, a group of faces, a video frame, the whole video.

This section is organized as follows: we present the first component, our proposed EmoNet network (Subsec. \ref{sec:EmoNet}, to extract visual representation for individual-level ER. Apply EmoNet into a video frame, we obtain a set of features for a group, where each feature represents an individual. The spatial relationship between these features, which is known as spatial group-level ER, is then modeled by Non-volume Preserving Fusion (NVPF) in our second component (Subsec .\ref{subsec:NVPF}). Finally, the temporal information between frames in a video is learned by TNVPF in our third component(Subsec. \ref{subsec:TNVPF})

\subsection{Individual-level ER via EmoNet}
\label{sec:EmoNet}

In this section, we propose a novel \textit{lightweight and high-performance} deep neural network design, named \textit{EmoNet}, to efficiently and accurately recognize individual-level emotion. Under the group-level ER problem, we observe that there is a large number of faces to be processed within one image, thus extracting their representations in feature space using a very deep network (e.g. Resnet101, DenseNet, etc.) could be very costly and ineffectively. 
Therefore, in our framework, we propose the EmoNet structure such that the information flow during the expression embedding process can be maximized while maintaining a relatively low computational cost. Our EmoNet designed structure is motivated by three main strategies: (1) performing convolutional operator faster and more efficient memory usage via depthwise separatable convolutional layers \citep{howard2017mobilenets}; (2) increasing the network capacity in embedding emotion features via bottleneck blocks with residual connections \citep{sandler2018mobilenetv2}; and (3) quickly reducing the spatial dimension in the first few layers while expanding the layers by depthwise.
Following those strategies, we propose the main architecture of our EmoNet containing convolutional layers, depthwise separable convolutional layers, a sequence of Bottleneck blocks with and without residual connections, and fully connected (FC) layers (see Table \ref{tb:network_structure} for more details).
The input of the EmoNet is a $112\times 112\times3$ face image that is cropped and aligned to remove unnecessary information for emotion recognition such as background, head hair, etc. 

A bottleneck block $\mathcal{B}$ in our EmoNet is composed of three main components: (1) a $1\times1$ convolution layer with ReLU activation - $\mathcal{B}_1$; (2) a $3\times3$ depthwise convolution layer with stride $s$ with ReLU activation - $\mathcal{B}_2$; and (3) a $1\times1$ convolution layer - $\mathcal{B}_3$. 
Given the input $\mathbf{x}$ having the size of $w\times h \times c$, the bottleneck block operator can be mathematically defined as
\begin{equation}
\centering
 \mathcal{B}(\mathbf{x}) = \left[ \mathcal{B}_3 \left( \mathcal{B}_2 \left( \mathcal{B}_1 \left(\mathbf{x} \right) \right) \right) \right]
\end{equation}
where $\mathcal{B}_1:\mathbb{R}^{w \times h \times c} \mapsto \mathbb{R}^{w \times h \times tc}$, $\mathcal{B}_2:\mathbb{R}^{w \times h \times tc} \mapsto \mathbb{R}^{\frac{w}{s} \times \frac{h}{s} \times tc}$ and $\mathcal{B}_3:\mathbb{R}^{w \times h \times c} \mapsto \mathbb{R}^{\frac{w}{s} \times \frac{h}{s} \times c_1}$. $t$ denotes the expansion factor.
The difference between the bottleneck block (BBlock) with and without residual connections is in the stride $s$. The stride $s$ is set to $1$ in BBlock with residual for learning residual features while it is set to $2$ in BBlock without residual to reduce the scale.

\subsection{Spatial Group-level ER via Non-volume Preserving Fusion (NVPF)}
\label{subsec:NVPF}

In this section, we present a novel fusion mechanism named Non-volume Preserving Fusion (NVPF), where a set of faces in a group is efficiently fused via a non-linear process with multiple-level CNN-based fusion units.

The end goal of this structure is to obtain a group-level feature in the form of probability density distributions for emotion recognition. 
In this way, rather than simply concatenating or applying the weighted linear combination, separated facial features of the subjects can be naturally embedded into a unified group-level feature in NVPF and, therefore, boosting the performance of emotion recognition in later steps.

Formally, 
given a set of $N$ faces $\{\textbf{f}_1, \textbf{f}_2, ... \textbf{f}_N\}$ of $N$ subjects in a group, we first extract their representations in latent space using the \textbf{EmoNet} structure as $ \textbf{x}_i = \text{EmoNet}(\textbf{f}_i) \in \mathbb{R}^M, i=1..N $. These features are then stacked into a grouped feature $\mathbf{S}$ as follows.
\begin{equation}
\small
    \mathbf{S} = \mathcal{G}(\textbf{x}_1, \textbf{x}_2, ..., \textbf{x}_N)
\end{equation}
where $\mathcal{G}$ denotes a grouping function.
Notably, there are many choices for $\mathcal{G}$, and stacking emotion features into a matrix $\mathbf{S} \in R^{M \times N}$ is among these choices. Any other choice can be easily adapted to this structure. Moreover, since the grouping operator $\mathcal{G}$ still treats $\textbf{x}_i$ independently, the direct usage of $\mathbf{S}$ for emotion recognition is equivalent to the trivial solution where no relationship between faces of a group is exploited. 
Therefore, to efficiently take this kind of relationship into account, we propose to model $\mathbf{S}$ in the form of density distributions in a higher-level feature domain $\mathcal{H}$. By this way, not only the feature $\textbf{x}_i$ is modeled, but also their relationship is naturally embedded in the distributions presented in $\mathcal{H}$. We define this mapping from feature domain $\mathcal{S}$ to a new feature domain $\mathcal{H}$ as the fusion process; and $\mathbf{S}$ and $\mathbf{H}$ can be considered as subject-level and group-level features, respectively.
Let $\mathcal{F}$ be a non-linear function that employs the mapping from $\mathcal{S} \in \mathrm{R}^{M \times N}$ to $\mathcal{H}\in \mathrm{R}^{M \times N}$.
\begin{equation}
\label{Eqn:Preprocessing}
\begin{split}
\mathcal{F}: &\mathcal{S} \rightarrow \mathcal{H}\\
& \mathbf{H} = \mathcal{F} (\mathbf{S};\mathbf{\theta}_{\mathcal{F}})
\end{split}
\end{equation}
The probability distribution of $\textbf{S}$ can be formulated by:

\begin{equation} \label{Eqn:change_variable}
    p_{\textbf{S}} (\textbf{S}; \mathbf{\theta}_\mathcal{F}) = p_{\textbf{H}} ( \textbf{H} ) \left| \frac{\partial \mathcal{F}(\textbf{S}; \mathbf{\theta}_\mathcal{F})}{\partial \textbf{S}} \right|
\end{equation}

Thanks to this formulation, computing the density function of $\textbf{S}$ is equivalent to estimate the density distribution of $\textbf{H}$ with an associated Jacobian matrix where it is triangular and its determinant can be efficiently computed without requiring to compute the Jacobians of the two features $\textbf{S}$ and $\textbf{H}$ \cite{Duong_2017_ICCV}.
By learning such a mapping function $\mathcal{F}$, we can employ a transformation from the subject-level feature $\textbf{S}$ to an embedding $\textbf{H}$ with a density $p_{\textbf{H}} ( \textbf{H} )$. This property brings us to the point such that if we consider  $p_{\textbf{H}} ( \textbf{H} )$ as a prior density distribution and choose the Gaussian Distribution for $p_{\textbf{H}} ( \textbf{H} )$,  $\mathcal{F}$ naturally becomes a mapping function from $\textbf{S}$ to a latent variable $\textbf{H}$ that distributed as a Gaussian.
Consequently, via $\mathcal{F}$, the subject-level feature can be fused into a unique Gaussian-distributed feature that embeds all information presented in each $\textbf{x}_i$ as well as among all  $\textbf{x}_i$ and $\textbf{x}_j$ in $\textbf{S}$.

To enforce non-linear property, we construct $\mathcal{F}$ as a composition of non-linear units $U_{\mathcal{F}_i}$ where each unit exploits a certain level of correlations (i.e. emotional similarity, connection, or interaction), between facial emotion features within a group of people.
\vspace{-3mm}
\begin{equation}
\mathcal{F}(\mathbf{S})  = \left( U_{\mathcal{F}_1} \circ U_{\mathcal{F}_2} \circ \cdots  \circ U_{\mathcal{F}_N} \right) (\mathbf{S}) 
\label{eqn:MappingFunc}
\end{equation} 
\vspace{-8mm}

As illustrated in Fig. \ref{fig:overall_network_architecture}, by representing $\textbf{S}$ as a feature map, convolutional operation is very effective in exploiting the spatial relationship between $\textbf{x}_i$ in $\textbf{S}$. Moreover, a longer-range relationship, i.e. $\textbf{x}_1$ vs. $\textbf{x}_N$ can be easily extracted by stacking multiple convolutional layers. Therefore, we propose to construct each mapping unit as a composition of multiple convolution layers. As a result, $\mathcal{F}$ becomes a deep CNN network with the capability of capturing  non-linear relationship embedded between faces in the group.
Notice that, different from other types of CNN networks, our NVPF network is formulated and optimized based on the likelihood of $p_{\textbf{S}} (\textbf{S}; \mathbf{\theta}_\mathcal{F})$ and the output is the fused group-level feature $\mathbf{H}$. 
Furthermore, to enable the easy-to-compute property of the determinant for each unit $U_{\mathcal{F}_i }$,
We adopt the structure of non-linear units in \citep{CDuong_ICCV2017} as follows.
\begin{equation} \label{eqn:maskedConvolution}
\textbf{Y} = (1 - \textbf{b}) \odot \left[ \mathcal{T}_1( \exp ( \mathbf{S}' ) ) + \mathcal{T}_2 ( \mathbf{S}' ) \right] + \mathbf{S}'
\end{equation}
where $\textbf{Y}$ is the output of the fusion unit $U_{\mathcal{F}_1 }$, $\mathbf{S}' = b \odot \mathbf{S}$, \textbf{b} is a binary mask where the first half of \textbf{b} is all one and the remaining is zero. $\odot$ denotes the Hadamard product. We adopt scale and the translation as the transformation $\mathcal{T}_1$ and $\mathcal{T}_2$, respectively.
In practice, the functions $\mathcal{T}_1$ and $\mathcal{T}_2$ can be implemented by a residual block with skip connections similar to the building block of Residual Networks (ResNet) \citep{he2016deep}. Then, by stacking fusion unit $U_{\mathcal{F}_i}$ together, the output $\textbf{Y}$ will be the input of the next fusion unit and so on. Finally, we have the mapping function as defined in Eqn. \eqref{eqn:MappingFunc}.

\textbf{Model Learning.} 
The parameters $ \mathbf{\theta}_{\mathcal{F}}$ of NVPF can be learned via maximizing the log-likelihood or minimizing the negative log-likelihood as follows.
\begin{equation} \label{Eqn:loss_ll}
\begin{split}
    \mathbf{\theta}^{\star}_{\mathcal{F}} &= \arg \min_{\mathbf{\theta}_{\mathcal{F}}} \mathcal{L}_{ll} =  
    - \log (p_{\textbf{S}} (\textbf{S})) 
    \\
    &=\arg \min_{\mathbf{\theta}_{\mathcal{F}}} - \log (p_{\textbf{H}} ( \textbf{H} )) - \log \left( \left| \frac{\partial \mathcal{F}(\textbf{S}, \mathbf{\theta}_\mathcal{F})}{\partial \textbf{S}} \right| \right)
\end{split}
\end{equation}
To further enhance the discriminative property of the features $\textbf{H}$, during the training process, we choose a different Gaussian distribution (i.e. different mean and standard deviation) for each emotion class. After optimizing the parameters $ \mathbf{\theta}_{\mathcal{F}}$, $\mathcal{F}$ has  capabilities of both transform subject-level features to group-level features and enforcing that feature to the corresponding distribution of the predicted emotion class. By matching the distribution, one can provide the emotion classification for the corresponding group-level feature.
For simplicity, we only consider the distribution of three classes, i.e. positive, negative and neutral, however, an arbitrary number of classes can be easily adopted by changing the class distribution.

\begin{figure*}[!t]
    \captionsetup{width=1.\linewidth}
	\centering \includegraphics[width=1.0\columnwidth]{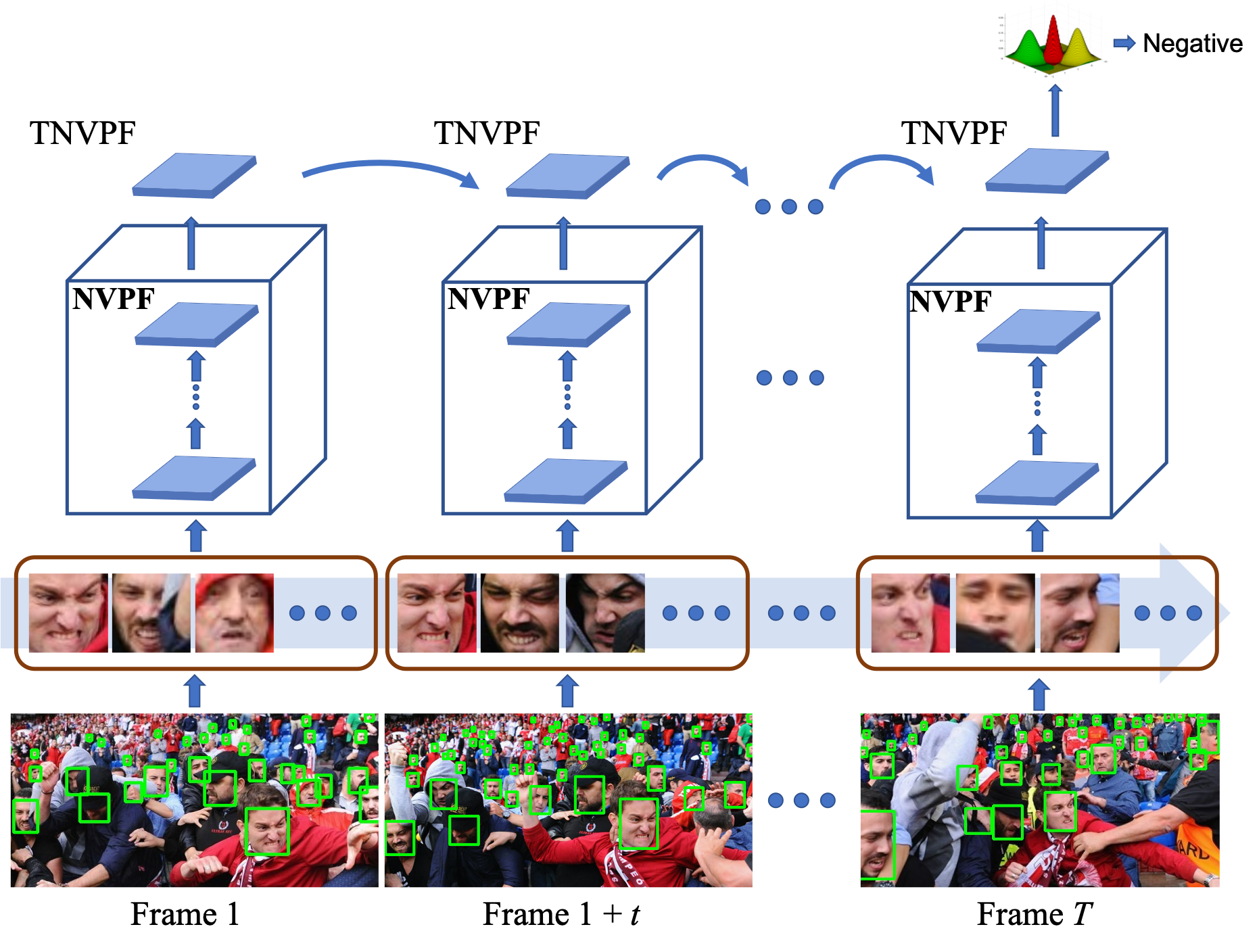}	
	\caption{Illustrated the proposed Temporal NVPF (TNVPF) ER framework on crowd videos.}
	\label{fig:framework_videos}
\end{figure*}

\subsection{Temporal-Spatial Group-level ER via Temporal Non-volume Preserving Fusion (TNVPF)}
\label{subsec:TNVPF}
In this section, we describe how to extend our proposed NVPF in sub-section \ref{subsec:NVPF} to a temporal-spatial fusion framework named Temporal NVPF (TNVPF) to handle videos instead of images while preserving temporal information from the input videos.
The main idea is to propagate the fused information from preceding frames. 

Inspired from Gated Recurrent Units (GRUs) \citep{Cho2014_GRU}, we design an end-to-end TNVPF framework with blocks of NVPF unit connected with memory and hidden units/states.
TNVPF structure is defined as.
\begin{equation} \label{Eqn:GRU_formulas}
    \begin{split}
        & \mathbf{o}_t = (1 - \mathbf{z}_t) \mathbf{o}_{t-1} + \mathbf{z}_t \tanh (\mathbf{W} \textbf{H}^t + \mathbf{U} (\mathbf{r}_t \odot \mathbf{o}_{t-1})) \\ \nonumber
        & \mathbf{z}_t = \sigma (\mathbf{W}_z \textbf{H}^t + \mathbf{U}_z \mathbf{o}_{t-1}) \\
        & \mathbf{r}_t = \sigma (\mathbf{W}_r \textbf{H}^t + \mathbf{U}_r \mathbf{o}_{t-1} ) \\
  \end{split}
\end{equation}
where $\textbf{U}$ is the input-to-hidden weight matrix, $\bf{W}$ is the state-to-state recurrent weight matrix. 
First, given the input $\mathbf{S}^t$ at frame $t$, all the fused group-level features $\mathbf{H}_g$ of group $g$-th ($g \in [1,G]$) in frame $t$ are stacked together. 
Then, the stacked features are fused at frame-level by another NVPF as $\textbf{H}^t = \mathcal{F}_G \left( \mathcal{G}^t \left( \textbf{H}_1, \textbf{H}_2, \cdots, \textbf{H}_G \right) \right)$, where $\mathcal{G}^t$ is stacking operator and $\mathcal{F}_G $ is a non-linear function defined similarly as $\mathcal{F}$ in eq. \eqref{Eqn:Preprocessing} Section \ref{subsec:NVPF}.
At time step $t$, each TNVPF unit takes $\textbf{H}^t$ and previous state $\textbf{o}_{t-1}$ as inputs and goes through a reset gate $\textbf{r}_t$ and an update gate $\textbf{z}_t$ to compute the next output $\textbf{o}_{t}$. 
In the end, TNVPF will output the predicted video-level emotion categories, i.e. positive, negative or neutral. Frame-level emotion prediction can also be obtained from the fused features from individual frames.  
Fig. \ref{fig:framework_videos} shows the overall TNVPF framework for group-level ER on videos.

TNVPF can be optimized via minimizing the negative log-likelihood of training sequences as.
\begin{equation}
    \mathbf{\theta}^{\star}_{\mathcal{F}}, \mathbf{\theta}^{\star}_{\mathcal{G}} = \min_{\mathbf{\theta}_{\mathcal{F}}, \mathbf{\theta}_{\mathcal{G}}} \mathcal{L}_{\mathcal{G}}( \mathbf{\theta}_{\mathcal{F}}, \mathbf{\theta}_{\mathcal{G}} )
\end{equation}
\begin{equation}
\begin{split}
    \mathcal{L}_{\mathcal{G}} ( \mathbf{\theta}_{\mathcal{F}}, \mathbf{\theta}_{\mathcal{G}} ) &= - \sum_t^T \left( \log (p (l_t | \textbf{S}^{1:t}; \mathbf{\theta}_{\mathcal{F}}, \mathbf{\theta}_{\mathcal{G}})) \right) p (l_t | \textbf{S}^{1:t}; \mathbf{\theta}_{\mathcal{F}}, \mathbf{\theta}_{\mathcal{G}}) \\ 
    &= \frac{e ^ {\mathbf{W}_h \mathbf{h}_t + \mathbf{b}_h}}{\sum_{c=1}^C e ^ {\mathbf{W}_h \mathbf{h}_t + \mathbf{b}_h}} \nonumber
\end{split}
\end{equation}
where $C$ is the class number ($C=3$). $\mathbf{\theta}_{\mathcal{F}}$ and $\mathbf{\theta}_{\mathcal{G}}$ are parameters of the TNVPF. $ l_t $ is the emotion label of the video frame $t$-th. $\mathbf{W}$ and $\mathbf{b}_h$ are the weight and bias for the hidden-to-output connections of TNVPF.

\subsection{Implementation Details}

\textbf{Data Preprocessing.} 
All faces are firstly detected using RetinaFace \citep{deng2020retinaface} and then aligned to a predefined template, i.e. based on five landmark points, using similarity transformation with a fixed size of $112 \times 112$. The five landmark points including eyes, nose, and mouth corners are given by RetinaFace \citep{deng2020retinaface} detector. Each cropped face can now go through EmoNet to obtain emotion features of the corresponding face and to provide individual ER output.
To further classify the emotion of a group of people, we first train a region proposal network (RPN) to provide clustered regions of faces. 
The details of the RPN will be described in the next section.

\begin{table}[!t]
	\small
	\centering
	\caption{The model architecture of EmoNet for facial feature extraction. Each row describes the configuration of a layer/block as input size, number of blocks (B), operators, stride (S), number of output channels (C), and residual connection (R).}
	\label{tb:network_structure} 
	\begin{tabular}{c|c|c|c|c|c}
		\hline
		\textbf{Input size} & \textbf{B} & \textbf{Operators} & \textbf{S} & \textbf{C} & \textbf{R} \\  \hline \hline
		112 $\times$ 112 $\times$ 3 & 1 & Conv 3$\times$3 & 2 & 64 & \xmark \\
		\hline
		56 $\times$ 56 $\times$ 64 & 1 & DWConv 3$\times$3 & 1 & 64 & \xmark \\
		\hline
		\multirow{ 3}{*}{56 $\times$ 56 $\times$ 64} & \multirow{ 3}{*}{2} & Conv 1$\times$1 & 1 & 128 & \multirow{ 3}{*}{\xmark} \\
                                     & & DWConv 3$\times$3 & 2 & 128 &  \\
                                     & & Conv 1$\times$1, Linear & 1 & 64 & \\
		\hline
		\multirow{ 3}{*}{28 $\times$ 28 $\times$ 64} & \multirow{ 3}{*}{4} & Conv 1$\times$1 & 1 & 128 & \multirow{ 3}{*}{\xmark}\\
                                     & & DWConv 3$\times$3 & 2 & 128 &  \\
                                     & & Conv 1$\times$1, Linear & 1 & 128 & \\
        \hline
		\multirow{ 3}{*}{14 $\times$ 14 $\times$ 128} & \multirow{ 3}{*}{2} & Conv 1$\times$1 & 1 & 256 & \multirow{ 3}{*}{\cmark}\\
                                     & & DWConv 3$\times$3 & 1 & 256 &  \\
                                     & & Conv 1$\times$1, Linear & 1 & 128 & \\
        \hline
		\multirow{ 3}{*}{14 $\times$ 14 $\times$ 128} & \multirow{ 3}{*}{4} & Conv 1$\times$1 & 1 & 256 & \multirow{ 3}{*}{\xmark}\\
                                     & & DWConv 3$\times$3 & 2 & 256 &  \\
                                     & & Conv 1$\times$1, Linear & 1 & 128 & \\
        \hline
		\multirow{ 3}{*}{7 $\times$ 7 $\times$ 128} & \multirow{ 3}{*}{2} & Conv 1$\times$1 & 1 & 256 & \multirow{ 3}{*}{\cmark}\\
                                     & & DWConv 3$\times$3 & 1 & 256 &  \\
                                     & & Conv 1$\times$1, Linear & 1 & 128 & \\
        \hline
        7 $\times$ 7 $\times$ 128 & 1 &  Conv 1$\times$1 & 1 & 512 & \xmark\\
        \hline
        7 $\times$ 7 $\times$ 512 & 1 & 512-d FC & -- & 512 & \xmark \\
        \hline
        1$\times$1 $\times$ 512 & 1 & M-d FC & -- & M & \xmark \\
        \hline
	\end{tabular}
\end{table}

\textbf{Network Architectures.}
Table \ref{tb:network_structure} shows the detailed architecture of the proposed \textbf{EmoNet}. Particularly, we have the first two convolution layers with $3 \times 3$ filters followed by a couple of bottleneck blocks with or without residual connection and then a convolution layer with $1 \times 1$ filters, two fully connected layers at the end. Each bottleneck block consists of a convolution layer with $1 \times 1$ filters followed by a convolution layer with $3 \times 3$ filters and then a convolution layer with $1 \times 1$ filters. Some convolution layers are depth-wise convolution layers (DWConv) and all the layers are followed by batch norm and Relu activation except those layers noted with ``linear", i.e. not using any action function. This design is proven to have good performance yet effective in terms of running time as demonstrated in another work \cite{duong2019mobiface} in facial recognition. 

To implement \textbf{RPN} for generating sub-window proposals, we use a similar structure of RPN in Faster-RCNN \citep{ren2015faster} to propose candidate sub-windows containing a group of faces.
The backbone architecture is ResNet-18 with only the convolutional layers being used to compute $512-d$ feature maps (the average pooling layer and the FC layer are removed). These feature maps are then used by RPN which consists of a $3 \times 3$ convolutional layer with ReLUs followed by two parallel $1 \times 1$ convolutional layers, i.e for box regression ($reg$) and class score ($cls$), respectively. RPN simultaneously predicts $k$ sub-window proposals at each location of the conv feature map. Then, the $reg$ layer will provide $4k$ outputs corresponding to the coordinates of $k$ sub-windows while the $cls$ layer will give $2k$ scores indicating the probability of face/non-face for each sub-window. Instead of directly predicting coordinates of $k$ sub-windows, we predict the parameters of $k$ sub-window proposals with respect to $k$ template sub-windows, referred to as anchors. At each feature map position, the template sub-window has a scale and aspect ratio, with 3 different scales and 3 aspect ratios gives us $a = 9$ anchors and $ W \times H \times a $ anchors in totals for a $W \times H$ feature map. We train the RPN with our collected database described in Sec. \ref{subsec:database} with a similar training procedure as in Faster-RCNN \citep{ren2015faster}. Our RPN has an mAP of 86.4\% on our validation set.

For each fusion unit \textbf{NVPF}, we use resnet-like \citep{He_2016_CVPR}  architecture to implement the non-linear mapping function $ \mathcal{F} $. 
Particularly, the non-linear mapping function $ \mathcal{F} $ has \textbf{10} fusion units $U_{\mathcal{F}_i}$.
Two transformations $\mathcal{T}_1$ and $\mathcal{T}_2$ in each fusion unit $U_{\mathcal{F}_i}$ are implemented by two residual network (ResNet) blocks with rectifier non-linearity and skip connections. The filter size of convolution layers is set to $3 \times 3$ and the number of filters/feature maps is set to 32. 
\textbf{TNVPF} has 4096 memory and hidden units. We first train TNVPF with two time-step then extend it further to five time-step. 

\textbf{Training and Testing Configurations.} 
In the training stage, the batch size is set to 512, 256, 64, 64 for training EmoNet, RPN, NVPF and TNVPF, respectively. The learning rate starts from 0.1 and the momentum is 0.9. We use Adam optimizer \citep{kingma2014adam} to train all the models.
All the models are trained in MXNET environment with a machine of Core i7-6850K @3.6GHz CPU, 64.00 GB RAM with four P6000 GPUs.
We ran inference on a single Nvidia GTX 1080Ti GPU machine and it took 8ms for face detection, 4ms for EmoNet to extract features on each face, and an average of 0.2s for NVPF to compute fused features for each frame/image and $\sim$0.5s for TNVPF to predict emotion class for the whole video and a total of $\sim$50s for a 10s full HD (1280 $\times$ 720) video.

\section{Our Collected GECV Datasets}
\label{subsec:database}
In this section, we introduce our newly collected database named Group-level Emotion on Crowded Videos (GECV) \footnote{GECV datasets are available at \url{https://bit.ly/3gnZA48}} to study ER at group-level in crowd videos. The presented GECV dataset contains 627 videos in total. Each video has about 300 frames ranging from 10 to 20 secs in duration and we found that empirically this is the average duration for a scene in videos. Each video frame consists of more than two people, which we define as the minimal number for a crowd, and it is determined based on the average number of detected faces across all video frames. To the best of our knowledge, the proposed GECV is the first video database that contains video footage and annotations for group-level ER on videos. The comparison between the properties of this database and others is presented in Table \ref{tb:dataset_compare}.
All videos have been collected by using search engines such as Google and YouTube to locate videos that may contain crowds as defined above. Search criteria such as festival, marching, wedding party, parade, funeral, game shows, sport, stadium, congress meeting, etc. were used to find candidate videos. To create diversity among videos, we translated the keywords into different languages to obtain videos from various places. All chosen videos have high quality i.e. more than 480p in resolutions. 
By processing and annotating those videos, we obtained three sub-sets of the GECV dataset, namely \textbf{GEVC-SingleImg}, \textbf{GEVC-GroupImg}, and \textbf{GEVC-GroupVid} in the following steps.

\begin{table*} [!t]
\centering
\caption{Properties of recent databases in facial expression recognition on images/videos of individual/group}
\label{tb:dataset_compare} 
\resizebox{1.01\textwidth}{!}{
\begin{tabular}{|c|c|c|c|c|c|c|}
\hline
 \textbf{Emotion Databases} & \textbf{Data Type} & \textbf{Group-type} & \textbf{No. Images/Videos} & \textbf{Condition} & \textbf{No. Emotion Classes} & \textbf{Annotation} \\
 \hline
 AffectNet 
 & Images & Individual & ~1.5M images & in-the-wild & 8 emotion categories & image \\
 EmotioNet 
 & Images & Individual & ~1M images & in-the-wild & 23 emotion categories & image\\
 EmotiW-Video 
 & Videos & Individual  & 1426 short videos ($<6$s) & in-the-wild & 7 emotion categories & video \\
 EmotiW-Group 
 & Images & Group ($>3$) & ~17K images & in-the-wild & 3 classes & image\\
 \hline
 \hline
 \textbf{GEVC-SingleImg} & \textbf{Images} & \textbf{Individual} & \textbf{900K images} & \textbf{in-the-wild} & 8 emotion categories & face \\
 \textbf{GEVC-GroupImg} & \textbf{Images} & \textbf{Group ($>2$)} & \textbf{438K group regions} & \textbf{in-the-wild} & 3 classes & image regions \\
 \textbf{GEVC-GroupVid} & \textbf{Videos} & \textbf{Group ($>2$)} & \textbf{627 short videos ($\sim20$s)} & \textbf{in-the-wild} & 3 classes & frame \\
 \hline
\end{tabular}
}
\end{table*}

First, to obtain \textbf{GEVC-SingleImg}, we extracted individual frames for each collected video and then run a face detector using RetinaFace \citep{deng2020retinaface} to detect faces. Since only faces with a minimum size of 100x100 pixels are chosen, we have about 900K faces from all video frames. 
Then we annotated emotion categories of those faces in this subset based on checking the activation pattern of Action Units (AUs) \citep{fabian2016emotionet} as shown in Table \ref{tab:AU_emotion}. If such AU activation patterns appear in the face image, it can be annotated with the corresponding emotional category among eight emotional categories. 
We used the AU recognition model in \citep{fabian2016emotionet} to recognize AUs in facial images. Then we obtained annotated subset \textbf{GEVC-SingleImg} containing 900K facial images and their annotated emotion categories. To ensure the accuracy of emotion categories, we further manually check the labels to have a clean set of 10K facial images for each emotional category.

\begin{table}[!t]
	\small
	\centering
	\caption{Listed here are the prototypical AUs observed in each basic emotion category \citep{fabian2016emotionet}.}
	\label{tab:AU_emotion} 
	\begin{tabular}{c|c}
	\hline
\textbf{Category} & \textbf{AUs} \\ \hline \hline
Happy & 12, 25 \\
Sad & 4, 15 \\
Fearful & 1, 4, 20, 25 \\
Angry & 4, 7, 24 \\
Surprised & 1, 2, 25, 26 \\
Disgusted & 9, 10, 17 \\
Awed & 1, 2, 5, 25 \\
Neutral & - \\
\hline
	\end{tabular}
\end{table}

Next, to obtain \textbf{GEVC-GroupImg}, we clustered detected faces in a video frame based on their locations, i.e. center of the detect boxes, using $k-$means clustering approach (empirically we set $k=10$) with center faces being chosen as in \cite{vassilvitskii2006k} and generated group of faces regions.
Then we gave our annotators the image with those initial sub-windows (as shown in Fig. \ref{fig:overall_network_architecture}) and ask them to manually adjust and annotate the labels, i.e. positive, negative or neutral, 
of each sub-windows. We have three annotators working on more than 140K video frames.
To save time, each annotator was given full video clips where they can easily copy annotation from the previous frame to the next frame and modify as necessary. 
For quality assurance, we divided overlapped sets among annotators so that there is a certain number of video frames annotated by at least two annotators. In this way, we can validate how well the annotation was and we obtained about 95\% matched on emotion categories and 0.9 of Intersection over Union (IoU) between corresponding sub-windows.
This gave us the second subset \textbf{GEVC-GroupImg} containing 438K cropped sub-images, the locations of those sub-images within a video frame and their annotated emotion class.

Finally, the subset \textbf{GEVC-GroupVid} was built based on the keywords and the content of the whole videos, we manually classified them into 204 positive videos, 202 negative videos, and 221 neutral videos.
We named this \textbf{GEVC-GroupVid} in which we have multiple videos annotated with either one of three emotion states: positive, negative or neutral. To simplify our annotation task, we choose to annotate only three categories for group-level emotion and leave it for future works to provide fine-grained emotions for groups of people.

\section{Experimental Results}
\label{sec:Exp}

In this section, we first introduce our newly collected GECV dataset for ER on crowd videos in sub-section \ref{subsec:database}. Then, the proposed EmoNet will be benchmarked and compared against other prior ER methods on AffectNet database in sub-section \ref{subsec:Exp_singleExp}. The proposed NVPF approach is evaluated and compared against established methods on EmotiW2017 and EmotiW2018 challenges in sub-section \ref{subsec:Exp_EmotiW2018}. Finally, our proposed TNVPF framework will be evaluated on some crowd videos GECV dataset in sub-section \ref{subsec:Exp_GroupVid}.

\subsection{Benchmarking the proposed EmoNet on Single Subject Emotion}
\label{subsec:Exp_singleExp}

\begin{figure}[!t]
    \centering
    \includegraphics[width=\columnwidth]{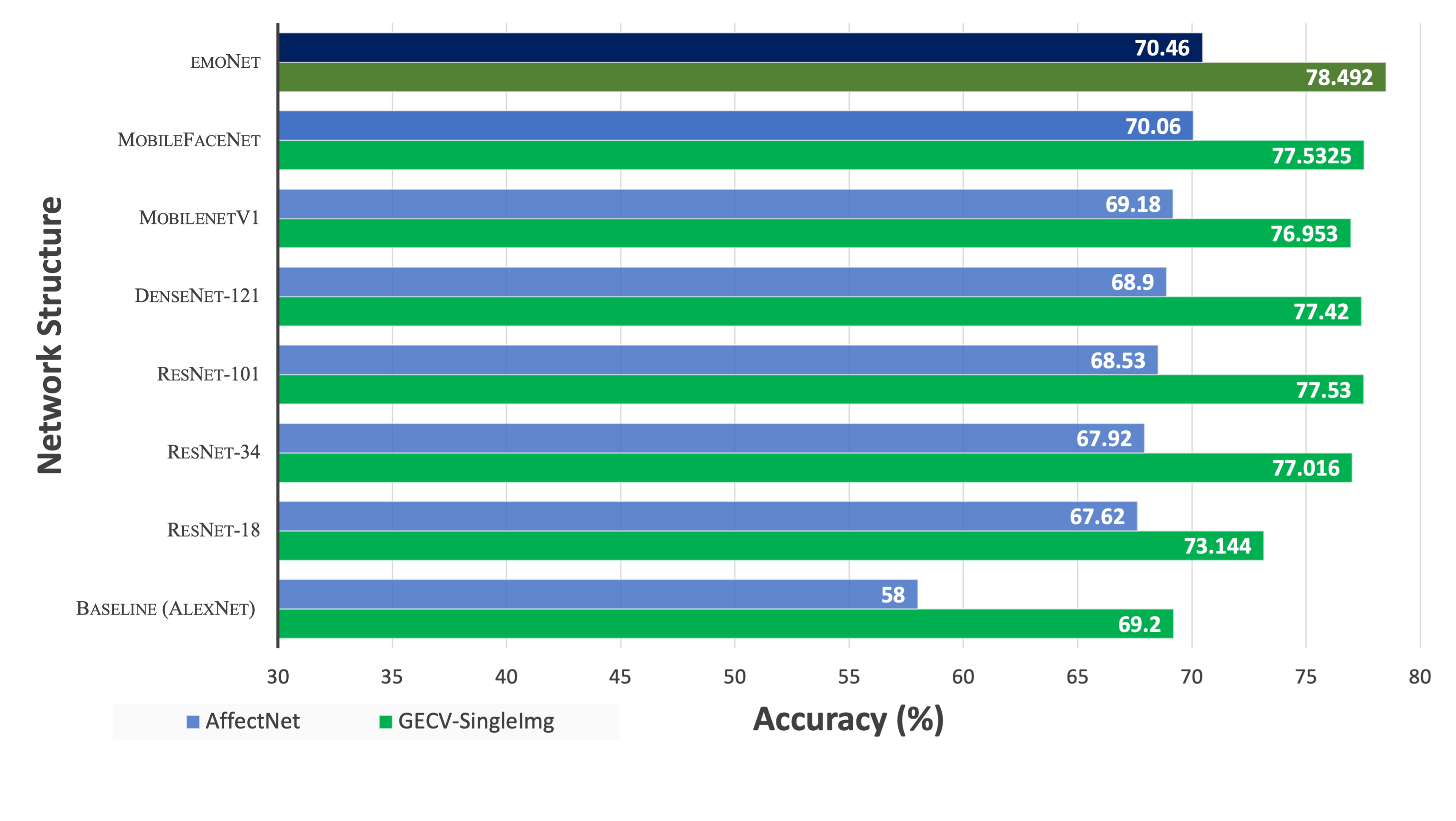}
    \caption{Compare the performance of our proposed network (EmoNet) against other networks on AffectNet and GECV-SingleImg dataset \citep{Mollahosseini2018affectnet}}
    \label{fig:network_comparison}
\end{figure}

To demonstrate the effectiveness of the proposed EmoNet on recognizing facial expression on a single object, we use AffectNet dataset \citep{Mollahosseini2018affectnet} to benchmark the proposed network and make a comparison against other state-of-the-art including: AlexNet (reported baseline) \citep{krizhevsky2012imagenet}, ResNet-18 \citep{he2016deep}, ResNet-34 \citep{he2016deep}, ResNet-101 \citep{he2016deep}, DenseNet-121 \citep{huang2017densely}, MobileNetV1 \citep{Howard-MobileNet}, MobileFaceNet \citep{chen2018MobileFaceNets}, etc.
AffectNet database is organized in such a way that there are 415,000 images for training and 5,500 images for validation. All the images are manually annotated with seven facial expression categories. However, the training set of this database is highly imbalanced, for example, the ``happy" class has about 100K images whereas some other classes like fear or disgust, only have few thousand images. 
Fig. \ref{fig:network_comparison} shows the performance of our proposed EmoNet compared against other networks on the AffectNet database. While EmoNet gives highly accurate recognizing emotion, its model size remains small ($<10$MB).
Similarly, we split our GECV-SingleImg into 64,000 images for training and 16,000 images for validation and compare our EmoNet with other networks on our GEVC-SingleImg validation set as shown in Fig. \ref{fig:network_comparison}.

\subsection{Benchmarking the proposed NVPF on Group-level Emotion on Crowd Images}
\label{subsec:Exp_EmotiW2018}

In this section, the group-level datasets from both EmotiW 2017 and 2018 challenges are used to benchmark the proposed NVPF fusion mechanism and compare against other recent works on group-level ER with different fusion strategies. EmotiW 2017 group-based ER sub-challenge contains 3,630 training, 2,068 validation, and 772 testing images. EmotiW 2018 group-based ER sub-challenge is an extension of the sub-challenge in EmotiW 2017 with 9,815 images for training, 4,346 images for validation, and 3,011 for testing, respectively. Meanwhile, EmotiW 2019 present a new sub-challenge, called group-level cohesion, which aims at predict group cohesion labels, i.e. the tendency for a group to be united for a common goal or emotion. Since our paper mainly focuses on group-level ER, we conduct our experiments on EmotiW 2017 and 2018. Although our proposed framework is applicable to group cohesion prediction in EmotiW 2019, we leave it as our future work. 
To evaluate only the proposed NVPF component and compare it against other fusion mechanisms, we have made various experiments on EmoNet (Sec.\ref{sec:EmoNet}.) using different fusion strategies including averaging score fusion, concatenating feature fusion, and NVPF. We name (i) Fused EmoNetA (FeA) for the framework where EmoNet is used for facial expression extracting together score fusion level with averaging mechanism; (ii)  Fused EmoNetB (FeB) for the framework where EmoNet is used for facial expression extracting together feature fusion level with concatenating mechanism; (iii) Fused EmoNetC (FeC) for the framework where EmoNet is used for facial expression extracting together feature fusion level with the proposed NVPF. The performance of three frameworks FeA, FeB, FeC are evaluated on EmotiW2018 challenge which is an extension of EmotiW2017 challenge. Overall/mean accuracy, per class accuracy, mean F1, and Unweighted Average Recall (UAR) are reported in this experiment.
Table \ref{tab:emotiW2018_results} summarizes all the state-of-the-art approaches on the EmotiW2017 and EmotiW2018 challenges and the performance of our model EmoNet with different fusion schemes (FeA, FeB, FeC) on the EmotiW 2018. 
As we can see from Table \ref{tab:emotiW2018_results}, our model using EmoNet network to extract features and NVPF scheme to fuse those features gives the best results among all other group-level ER approaches on the EmotiW2018. These results are also consistent with our ablation study on our GECV-GroupImg validation set for different types of fusion using features from our EmoNet model in Table \ref{tab:gecv_group_results}.

\begin{table*} [!h]
	\small
	\caption{The results of predicting labels in the \textbf{validation set} on EmotiW2017 \& EmotiW2018 for different fusion approaches on mean accuracy (mAC), unweighted average recall (UAR), F1-score, and class accuracy.} 
	\label{tab:emotiW2018_results}
	\centering
	\resizebox{1.01\textwidth}{!}{
    \begin{tabular}{|c|c|p{1.4cm}|p{1.2cm}|p{1cm}|c|c|c|c|c|c|}
    \hline
     \textbf{Model} & \textbf{EmotiW} & \textbf{Network / Feature} &\textbf{Fusion scheme} & \textbf{Fusion Stage} & \textbf{mAC} & \textbf{UAR} & \textbf{F1} & \textbf{Neu} & \textbf{Pos} & \textbf{Neg} \\
     \hline
     Dhall et al. \citeyear{dhall2015more} & 2017 & CENTRIST & Kernel & Feature & 51.47\% & -- & -- & 63.95\% & 38.33\% & 46.55\% \\
     \hline
     Tan et al. \citeyear{tan2017group} & 2017 & SphereFace & Averaging & Score & 74.1\% & -- & -- & -- & -- & -- \\ \hline
     Wei et al. \citeyear{wei2017new} & 2017 & VGG-Face & LSTM & Feature & 74.14 \% & -- & -- & -- & -- & -- \\ \hline
     Rassadin et al. \citeyear{rassadin2017group} & 2017 & VGG-Face & Median & Feature & 70.11 \% & -- & -- & -- & -- & -- \\ 
     \hline
     \hline
     Khan et al. \citeyear{Khan_2018} & 2018 & ResNet18 & Averaging & Score & 69.72\% & -- & -- & -- & -- & -- \\ \hline
     Gupta et al \citeyear{Gupta2018attention} & 2018 & SphereFace & Averaging & Feature & 73.03\% & -- & -- & -- & -- & -- \\ \hline
     Gupta et al. \citeyear{Gupta2018attention} & 2018 & SphereFace & Attention & Feature & 74.38\% & -- & -- & -- & -- & -- \\ \hline
     Guo et al. \citeyear{guo2017group} & 2018 & VGG-Face & Concat & Feature & 74\% & 0.74 & -- & 66.48\% & 75.68\% & 81.38\%\\
     \hline
     \hline
     Our FeA & 2018 & EmoNet & Averaging  & Score & 73.7\% & 0.733 & 0.7322 & 61\% & 87\% & 72\% \\ \hline 
     Our FeB & 2018 & EmoNet & Concat  & Feature & 72.88\% & 0.7104 & 0.7125  & 83\% & 78\% & 52\% \\ \hline 
     \textbf{Our FeC} & \textbf{2018} & \textbf{EmoNet} & \textbf{NVPF} & \textbf{Feature} & \textbf{76.12\%} & \textbf{0.7418} & \textbf{0.7381}  & \textbf{84\%} & \textbf{88\%} & \textbf{82\%} \\ 
     \hline
\end{tabular}
}
\end{table*}

\begin{table} [!h]
\small
	\caption{The results of predicting labels in our GECV-GroupImg validation set for different fusion approaches on mean accuracy (mAC), unweighted average recall (UAR), F1-score, and class accuracy.}
	\label{tab:gecv_group_results}
	\centering
	\resizebox{1.0\textwidth}{!}{
    \begin{tabular}{|c|c|p{1.4cm}|p{1.2cm}|p{1cm}|c|c|c|}
    \hline
     \textbf{Model} & \textbf{Network / Feature} &\textbf{Fusion scheme} & \textbf{Fusion Stage} & \textbf{mAC} & \textbf{UAR} & \textbf{F1} \\
     \hline
     Our FeA & EmoNet & Averaging  & Score & 74.8\% & 0.724 & 0.7358 \\ \hline 
     Our FeB & EmoNet & Concat  & Feature & 73.5\% & 0.688 & 0.7107 \\ \hline 
     \textbf{Our FeC} & \textbf{EmoNet} & \textbf{NVPF} & \textbf{Feature} & \textbf{77.02\%} & \textbf{0.748} & \textbf{0.7588} \\ 
     \hline
\end{tabular}
}
\end{table}

\subsection{Benchmarking the proposed TNVPF on Group-level Emotion on Crowd Videos}
\label{subsec:Exp_GroupVid}

\begin{figure}  [!h]
    \centering
    \includegraphics[width=1\columnwidth]{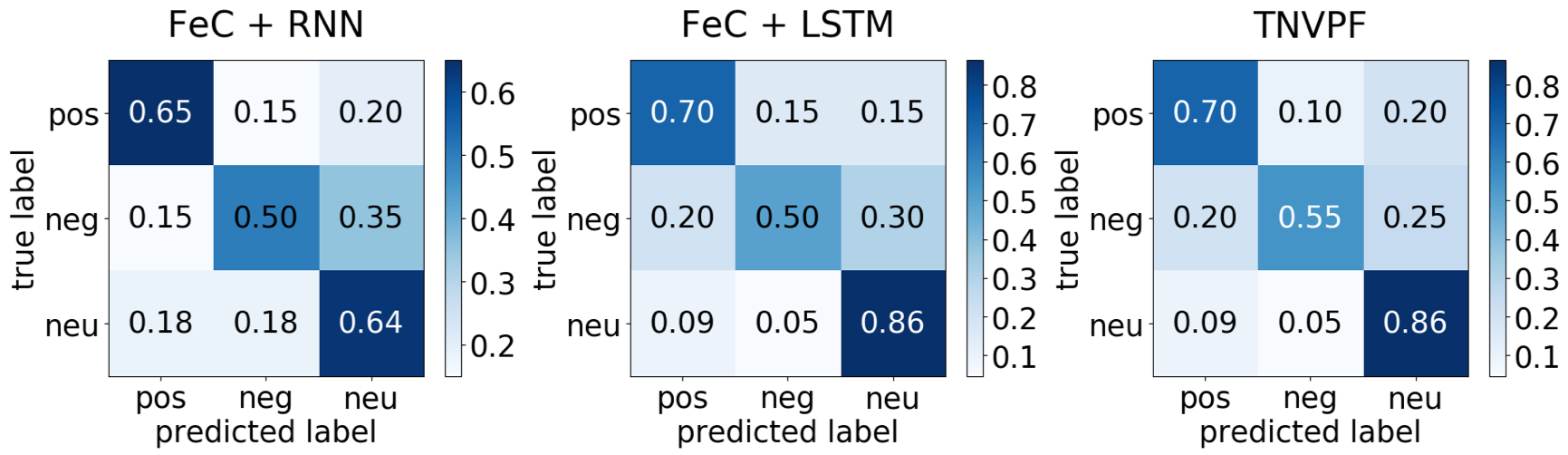}
    \caption{Confusion matrices of our proposed framework
    on our GECV-GroupVid dataset.}
    \label{fig:confusion_matrices}
\end{figure}

In this section, we use our presented GECV dataset to benchmark the proposed TNVPF for recognizing group-level emotion on crowd videos. The GECV dataset contains 627 crowd videos which partitions into 90\% for training and 10\% for testing (565 videos for training and 62 videos for testing. In addition to the achievement of the proposed TNVPF, we also examine the performance of NVPF on another temporal modeling such as Vanilla RNNs, Long Short Term Memory (LSTM). As shown in the previous experiment, FeC by the proposed NVPF fusion mechanism gives the best performance on EmotiW; thus, we choose FeC for further evaluation in this section. Table \ref{tab:video-based_methods} shows the performance of FeC on different temporal models (FeC+RNNs, FeC+LSTM, and TNVPF) whereas the experiment on the proposed TNVPF built upon FeC model and a derivation of GRUs obtains the best performance. Fig. \ref{fig:confusion_matrices} illustrates the confusion matrices of those three approaches (FeC+RNNs, FeC+LSTM, and TNVPF).

\begin{table} [!h]
	\small
	\centering
	\caption{The results on our GECV-GroupVid dataset of group-level ER on mean accuracy (mAC) and accuracy per class.}
	\label{tab:video-based_methods}
    \begin{tabular}{|c|c|c|c|c|c|c|}
    \hline
     \textbf{Method} &  \textbf{mAC} & \textbf{Pos} & \textbf{Neg} & \textbf{Neu} \\
     \hline
     \hline
    FeC + RNN  & 59.68\% & 65\% & 50\% & 63.64\% \\ 
    FeC + LSTM  & 69.35\% & 70\% & 50\% & 86.36\% \\
    \textbf{TNVPF} & \textbf{70.97\%} & \textbf{70\%} & \textbf{55\%} & \textbf{86.36\%} \\ 
     \hline
    \end{tabular}

\end{table}

\section{Conclusion}

This paper has first presented a high-performance and low computation network named EmoNet for robustly extracting facial expression features. Then, a new fusion mechanism NVPF is proposed to deal with group-level emotion in crowds where multiple emotions may occur within a frame and human faces are not always clearly identified, e.g. large sports scenes where by nature their faces are shown in low resolution. The proposed NVPF is extended to TNVPF to model the temporal information between frames in crowd videos. 
Various experiments on emotion recognition have been conducted at three levels, i.e. individual, group of people, and the entire video, together with three sub-sets of our collected dataset, i.e. GEVC-SingleImg, GEVC-GroupImg, and GEVC-GroupVid. The experiments have demonstrated the robustness and effectiveness of each component of our proposed framework including the proposed EmoNet, NVPF, and TNVPF on public datasets (including AffectNet and EmotiW2018) as well as our collected dataset. We hope that our new dataset will foster future work in predicting group-level emotion in video and developing more large-scale datasets for the research community. Due to the limitation of face detectors, people counting and detecting in a crowd will potentially provide more useful features that can be extracted for crowd scene understanding.

\section*{Acknowledgments}
This material is based upon work supported in part by the US National Science Foundation, under Award No. OIA-1946391, NSF 1920920, NSF Data Science, Data Analytics that are Robust and Trusted (DART) and NSF WVAR-CRESH Grant.
The authors would like to thank the reviewers for their valuable comments.

\section*{Disclaimer}
Any opinions, findings, and conclusions or recommendations expressed in this material are those of the author(s) and do not necessarily reflect the views of the National Science Foundation.

\bibliographystyle{model2-names}
\bibliography{refs}

\clearpage

\section*{Author Biographies}
\parpic{\includegraphics[width=1in,clip,keepaspectratio]{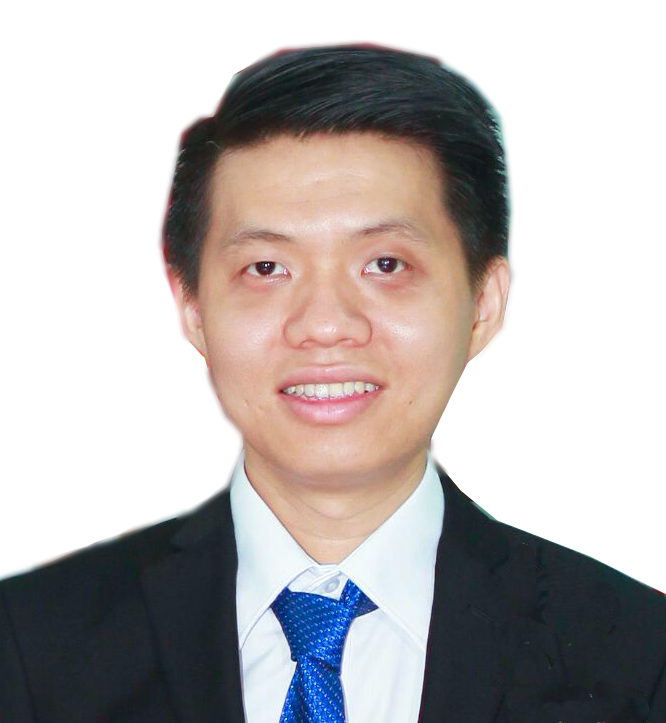}}
\noindent {\bf Kha Gia Quach} is currently a Senior Technical Staff while having research collaboration with both Computer Vision and Image Understanding (CVIU) Lab, University of Arkansas, USA, and Concordia University, Montreal, Canada. He has been a research associate in Cylab Biometrics Center at Carnegie Mellon University (CMU), USA since 2016. He received his Ph.D. degree in Computer Science under supervision of Prof. Tien Dai Bui and Prof. Khoa Luu from the Department of Computer Science and Software Engineering, Concordia University, Montreal, Canada, in 2018. He received his B.S. and M.Sc. degree in Computer Science from the University of Science, Ho Chi Minh City, Vietnam, in 2008 and 2012, respectively. His research interests lie primarily in Compressed Sensing, Sparse Representation, Image Processing, Machine Learning, and Computer Vision.
He has published and co-authored over 30 papers in top conferences including CVPR, BTAS, ICPR, ICASSP, and premier journals including TIP, PR, CVIU, IJCV, CJRS. 
He has been a reviewer of several top-tier journals and conferences including TPAMI, TIP, SP, CVPR, ICCV, ECCV, AAAI, MICCAI. He also served as a Program Committee Member of Precognition: Seeing through the Future, CVPR 2019, 2020 and 2021.

\parpic{\includegraphics[width=1in,clip,keepaspectratio]{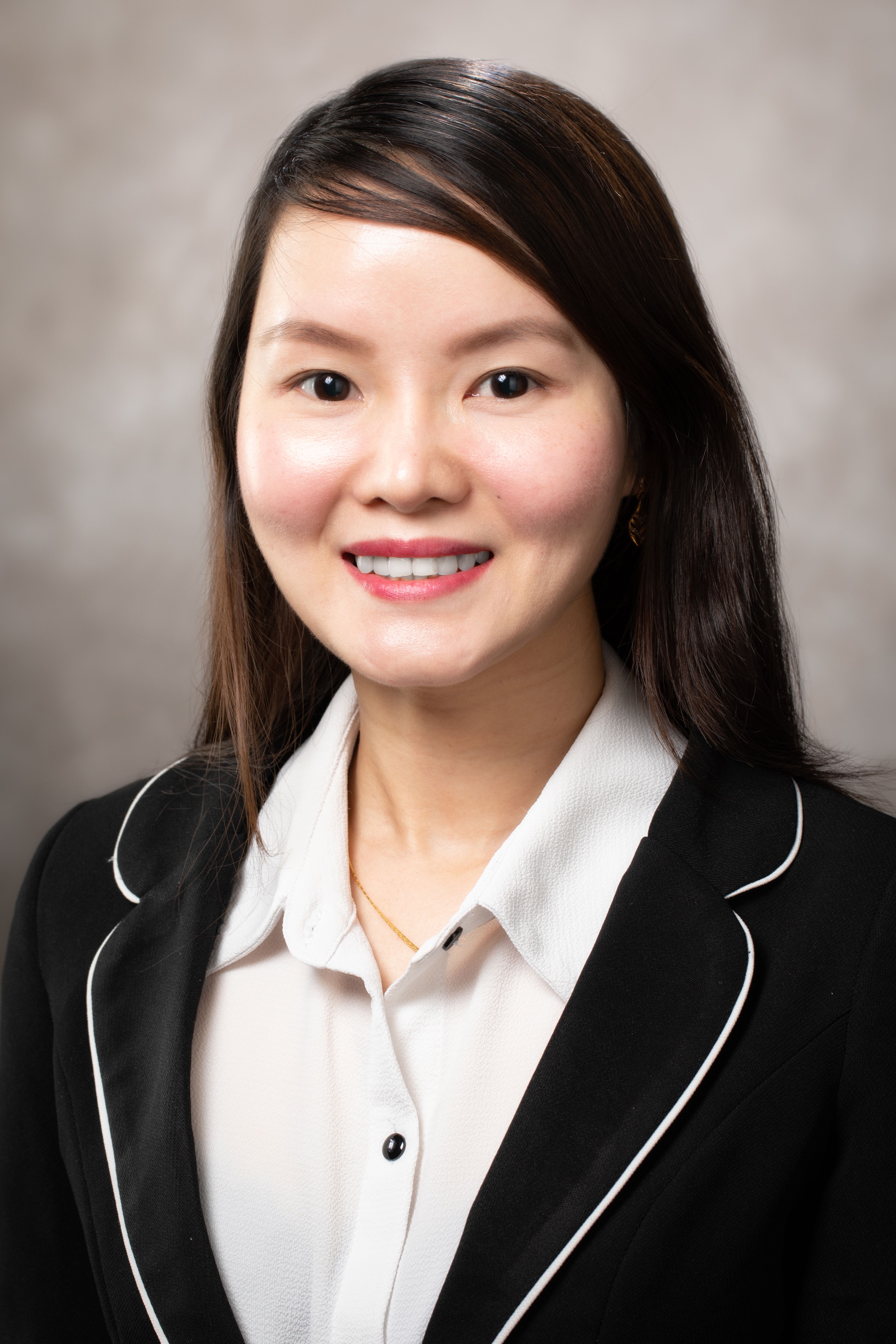}} 
\noindent \textbf{Ngan Le} Dr. Le is the director of Artificial Intelligence and Computer Vision (AICV) Lab and an Assistant Professor in the Department of Computer Science \& Computer Engineering at University of Arkansas. She was a research associate in the Department of Electrical and Computer Engineering (ECE) at Carnegie Mellon University (CMU) in 2018-2019. She received the Ph.D degree in ECE at CMU in 2018, ECE Master degree at CMU in 2015, CS Master Degree at University of Science, Vietnam in 2009 and CS Bachelor degree at University of Science, Vietnam in 2005.
Her current research interests focus on Image Understanding, Video Understanding, Computer Vision, Robotics, Artificial Intelligence (Machine Learning, Deep Learning, Reinforcement Learning), Biomedical Imaging.
Dr. Le is currently an Associate Editor of Elsevier Machine Learning with Applications, a Guest Editor of Scene Understanding in Autonomous (Frontier) and a Guest Editor of Artificial intelligence in Biomedicine and Healthcare (MDPI). She co-organized the Deep Reinforcement Learning Tutorial for Medical Imaging at MICCAI 2018, Medical Image Learning with Less Labels and Imperfect Data workshop at MICCAI 2019, 2020, 2021, Visual Detection, Recognition and Prediction at Altitude and Range at ICCV 2022. Dr. Le is instructor lead of the Google Machine Learning Bootcamp 2021. Her publications appear in the top-tier conferences including CVPR, MICCAI, ICCV, SPIE, IJCV, ICIP etc, and premier journals including IJCV, JESA, TIP, PR, JDSP, TIFS, etc. She has co-authored 72+ journals, conference papers, and book chapters, 9+ patents and inventions. She has served as a reviewer for 10+ top-tier conferences and journals, including TPAMI, AAAI, CVPR, NIPS, ICCV, ECCV, MICCAI, TIP, PR, TAI, IVC, etc.

\parpic{\includegraphics[width=1in,clip,keepaspectratio]{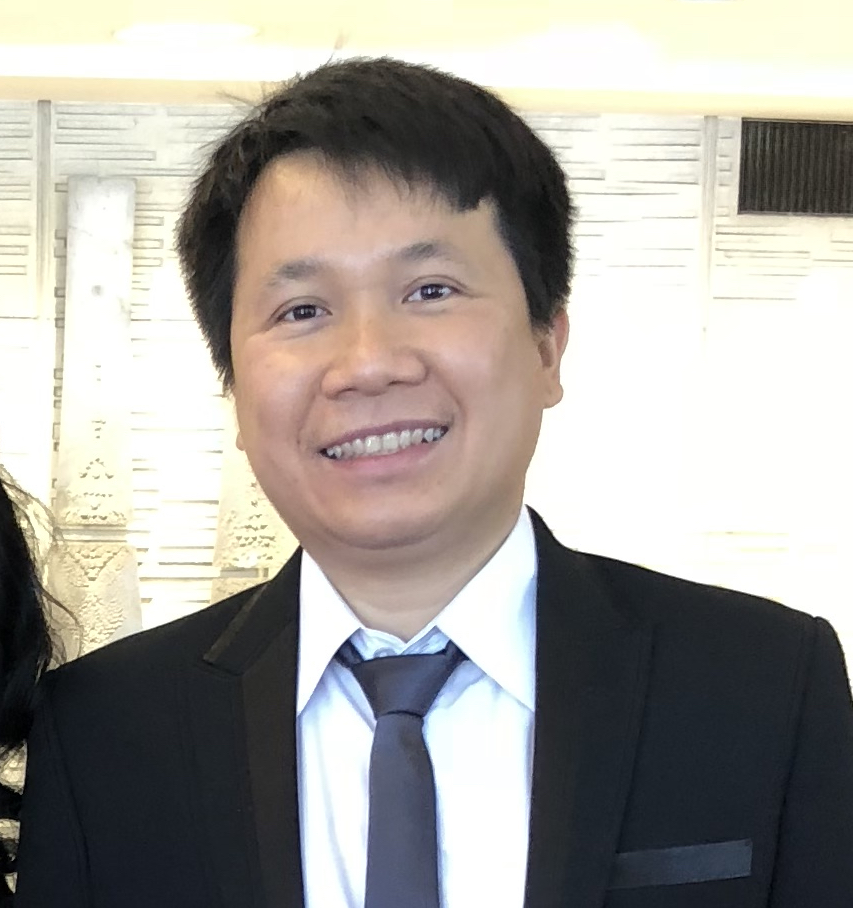}}
\noindent \textbf{Chi Nhan Duong} is currently a Senior Technical Staff and having research collaborations with both Computer Vision and Image Understanding (CVIU) Lab, University of Arkansas, USA, and Concordia University, Montreal, Canada.
He had been a Research Associate in Cylab Biometrics Center at Carnegie Mellon University (CMU), USA since September 2016. He received his Ph.D. degree in Computer Science under supervision of Prof. Tien Dai Bui and Prof. Khoa Luu from the Department of Computer Science and Software Engineering, Concordia University, Montreal, Canada. He was an Intern with National Institute of Informatics, Tokyo Japan in 2012. He received his B.S. and M.Sc. degrees in Computer Science from the Department of Computer Science, Faculty of Information Technology, University of Science, Ho Chi Minh City, Vietnam, in 2008 and 2012, respectively.    
His research interests include Deep Generative Models, Face Recognition in surveillance environments, Face Aging in images and videos, Biometrics, and Digital Image Processing, and Digital Image Processing (denoising, inpainting, and super-resolution).
He has published and co-authored over 30 papers in top-tier  conferences including CVPR, ICCV, BTAS, ICPR, ICASSP, and premier journals including IJCV, TIP, PR, CVIU.
He is currently a reviewer of several top-tier journals  including TPAMI, TIP, SP, PR, PR Letters, IEEE Access Trans. He is also recognized as an outstanding reviewer of several top-tier conferences such as CVPR, ICCV, ECCV, ICLR, AAAI. He is also a Program Committee Member of Precognition: Seeing through the Future, CVPR 2019, 2020, and 2021.

\parpic{\includegraphics[width=1in,clip,keepaspectratio]{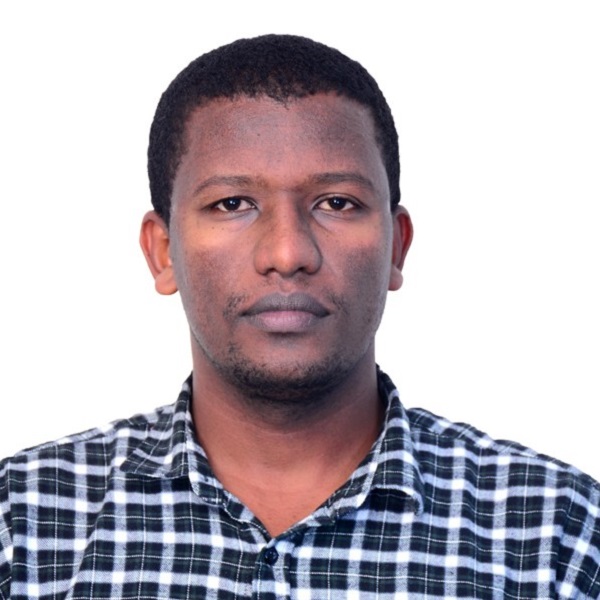}}
\noindent \textbf{Ibsa Jalata}
is a Ph.D. student under supervision of Prof. Khoa Luu in CVIU Lab at the University of Arkansas, in Arkansas. He received his MSc degree from Chonbuk National University in South Korea in 2017. He completed his BSc from Addis Ababa University in Ethiopia. He has been working as a Lecturer in Adama Science and Technology University, Ethiopia. His research interests include Activity Recognition, Action Localization, Face Detection and Recognition, and Biomedical Image Processing. 

\parpic{\includegraphics[width=1in,clip,keepaspectratio]{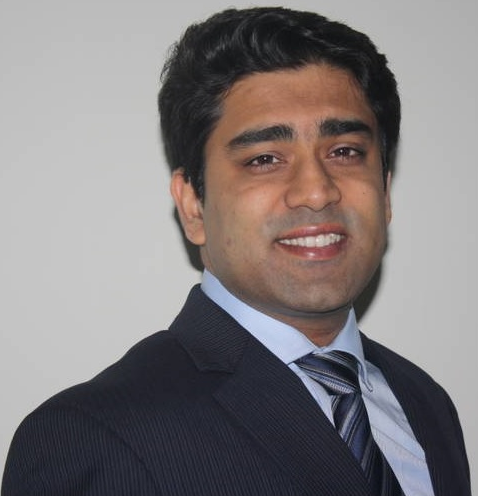}}
\noindent \textbf{Kaushik Roy}
is an Associate Professor in the Department of Computer Science at North Carolina A\&T State University (NCAT). He received his Ph.D. from Concordia University, Montreal, QC, Canada in 2011 in Computer Science. He completed his MS degree in Computer Science from the Concordia University in 2006 and B.Sc. degree in Computer Science \& Technology from University of Rajshahi, Bangladesh in 2000. Previously, he worked as a postdoctoral fellow in the Department of Electrical and Computer Engineering at University of Waterloo, ON, Canada during 2011-2012.
His current research is heavily focused on cybersecurity, cyber identity, biometrics, machine learning (deep learning), and big data analytics. He has over 140 publications, including 33 journal articles. He is the director of the Center for Cyber Defense at NCAT. He also leads the Cyber Identity and Biometrics (CIB) and Artificial Intelligence and Machine Learning Research (AiM) labs. His research is supported by the National Science Foundation (NSF), Department of Defense (DoD), Department of Energy, and CISCO Systems.

\parpic{\includegraphics[width=1in,clip,keepaspectratio]{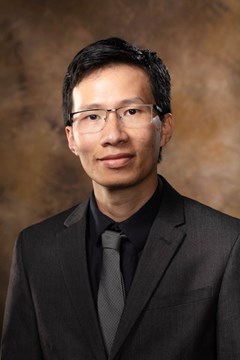}}
\noindent \textbf{Khoa Luu} Dr. Luu is currently an Assistant Professor and the Director of Computer Vision and Image Understanding (CVIU) Lab in Department of Computer Science \& Computer Engineering at University of Arkansas, Fayetteville. 
He was the Research Project Director in Cylab Biometrics Center at Carnegie Mellon University (CMU), USA. He has received four patents and two best paper awards and coauthored 100+ papers in conferences and journals. He was a vice-chair of Montreal Chapter IEEE SMCS in Canada from September 2009 to March 2011.
He is teaching Computer Vision, Image Processing and Introduction to Artificial Intelligence courses in 
CSCE Department at University of Arkansas, Fayetteville.
His research interests focus on various topics, including Biometrics, Image Processing, Computer Vision, Machine Learning, Deep Learning, Multifactor Analysis and Compressed Sensing. 
He is an co-organizer and a chair of CVPR Precognition Workshop in 2019, 2020, 2021; MICCAI Workshop in 2019, 2020 and ICCV Workshop in 2021. He is a PC member of AAAI, ICPRAI in 2020 and 2021.
He is currently a reviewer for several top-tier conferences and journals, such as CVPR, ICCV, ECCV, NeurIPS, ICLR, FG, BTAS, IEEE-TPAMI, IEEE-TIP, Journal of Pattern Recognition, Journal of Image and Vision Computing, Journal of Signal Processing, Journal of Intelligence Review, IEEE Access Trans., etc.
\end{document}